\documentclass[11pt,a4paper]{article}
\usepackage[hyperref]{acl2021}
\usepackage{times}
\usepackage{latexsym}
\usepackage{todonotes}
\usepackage{array,ragged2e}

\usepackage{microtype}
\usepackage{amsmath}
\DeclareMathOperator*{\argmax}{arg\,max}

\aclfinalcopy 

\title{UPB at SemEval-2021 Task 5: Virtual Adversarial Training for Toxic Spans Detection}

\author{Andrei Paraschiv, Dumitru-Clementin Cercel, Mihai Dascalu\\
   University Politehnica of Bucharest, Faculty of Automatic Control and Computers\\
  {\tt andrei.paraschiv74@stud.acs.upb.ro} \\
  \tt \{dumitru.cercel, mihai.dascalu\}@upb.ro}
  
\date{}

\begin{document}
\maketitle
\begin{abstract}
The real-world impact of polarization and toxicity in the online sphere marked the end of 2020 and the beginning of this year in a negative way. Semeval-2021, Task 5 - Toxic Spans Detection is based on a novel annotation of a subset of the Jigsaw Unintended Bias dataset and is the first language toxicity detection task dedicated to identifying the toxicity-level spans. For this task, participants had to automatically detect character spans in short comments that render the message as toxic. Our model considers applying Virtual Adversarial Training in a semi-supervised setting during the fine-tuning process of several Transformer-based models (i.e., BERT and RoBERTa), in combination with Conditional Random Fields. Our approach leads to performance improvements and more robust models, enabling us to achieve an F1-score of 65.73\% in the official submission and an F1-score of 66.13\% after further tuning during post-evaluation.
\end{abstract}

\section{Introduction}

Nowadays, online engagement in social activities is at its highest levels. The lockdowns during the 2020 COVID-19 pandemic increased the overall time spent online. In Germany for instance, \citet{lemenager2021covid} observed that 71\% of considered subjects increased their online media consumption during this period. Unfortunately, online toxicity is present in a large part of the social and news media platforms. As such, automated early detection is necessary since toxic behavior is often contagious and leads to a spillover effect \cite{kwon2017offensive}. 

Recently, a significant effort was put into the detection of toxic and offensive language \cite{van2018challenges, paraschiv2019upb, tanase2020detecting, tanase2020upb}, but the challenging nature of these problems leaves several avenues unexplored. In addition, most shared tasks focus on the distinction between toxic/non-toxic \cite{wulczyn2017ex, van2018challenges, juuti2020little} or offensive/non-offensive posts in various languages~\cite{struss2019overview, zampieri2019predicting, zampieri2019semeval,  zampieri2020semeval,  mandl2020overview, aragon2020overview}. The Semeval-2021 Task 5, namely Toxic Spans Detection \cite{pav2020semeval}, tackles the problem of identifying the exact portion of the document that gives it toxicity. The provided dataset is a subset of the Jigsaw Unintended Bias in Toxicity Classification dataset\footnote{https://www.kaggle.com/c/jigsaw-unintended-bias-in-toxicity-classification}, with annotated spans that represent toxicity from a document.

In this paper, we describe our participation in the aforementioned Toxic Spans Detection task using several Transformer-based models \cite{vaswani2017attention}, including BERT \cite{devlin2018bert} and RoBERTa \cite{liu2019roberta}, with a Conditional Random Field (CRF)  \cite{lafferty2001conditional} layer on top to identify spans that include toxic language. We introduce Virtual Adversarial Training (VAT) \cite{miyato2015distributional} in our training pipeline to increase the robustness of our models. Furthermore, we enhance part of our models with character embeddings based on the Jigsaw Unintended Bias dataset to improve their performance. Finally, we compare the proposed models and analyze the impact of various hyperparameters on their performance. 

The rest of the paper is structured as follows. The next section introduces a review of methods related to toxic language detection, sequence labeling, and adversarial training \cite{kurakin2016adversarial}. The third section discusses the employed models, as well as the VAT procedure. Results are presented in the fourth section, followed by discussions, conclusions, and an outline of possible future works.

\section{Related Work}

\textbf{Toxic Language Detection.}
There are several research efforts to detect toxic texts based on the Jigsaw Unintended Bias dataset, out of which most focus on the Kaggle competition task - predicting the toxicity score for a document. \citet{morzhov2020avoiding} compared models based on Convolutional Neural Networks (CNNs) \cite{kim2014convolutional} and Recurrent Neural Networks \cite{cho2014learning} with a Bidirectional Encoder Representations from Transformers (BERT) architecture \cite{devlin2018bert}, obtaining the best performance from an ensemble of all used models. \citet{gencoglu2020cyberbullying} and \citet{richard2020generalisation} used the same dataset to improve on the automatic detection of cyberbullying content.

\textbf{Sequence Labeling.}
Predicting the type for each token from a document rather than providing a label for the whole sequence is a task often associated with named entity recognition \cite{ma2016end}, but can be performed in other Natural Language Processing pipelines, including part-of-speech tagging \cite{ling2015finding} and chunking \cite{hashimoto2016joint}. A common practice in sequence tagging models \citep{peters2018deep, avram2020upb, ionescu2020upb} is to use a CRF as a final decoding layer.

\textbf{Adversarial Training.} Researched first in image classification \cite{szegedy2013intriguing}, adversarial examples are small input perturbations that are hardly distinguishable for humans, but can dramatically shift the output of a neural network. These examples can be used in adversarial training (AT) \cite{goodfellow2014explaining} as a regularization method that can increase the robustness of the model. Using the worst-case outcome from a distribution of small norm perturbations around an existing training sample, a new data point is created and inserted into the training process. 

Extending AT to a semi-supervised setting, VAT \citep{miyato2016adversarial} does not require label information for the adversarial examples. VAT aims to increase the local distributional smoothness by adding perturbations to the embedding output. Recently, several studies \citep{kumar2020nutcracker, liu2020kk2018, li2020tavat} focused on applying VAT in Transformer-based models and obtained improvements in comparison to baseline methods on several classification tasks.

\section{Method}

\subsection{Corpus}

\begin{figure}[htp]
    \centering
    \includegraphics[width=7.8cm]{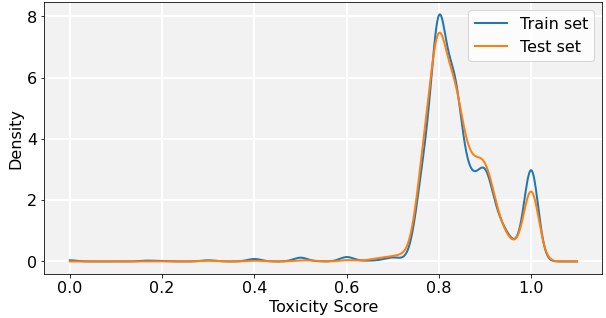}
    \caption{Kernel density estimate using Gaussian kernels for the toxicity scores in both the training and test data.}
    \label{fig:score}
\end{figure}

The dataset for the competition is a subset of the Jigsaw Unintended Bias in Toxicity Classification English language corpus, with annotated spans that make the utterance toxic. From the 8,597 trial and train records, 8,101 had at least one toxic span. By cross-referencing with the original Jigsaw dataset which contains additional information, we retrieved the toxicity scores for each text and determined that the mean toxicity score for the train and test set were very close (0.8429 versus 0.8440; see Figure \ref{fig:score} for corresponding kernel density estimates). Moreover, only 17 out of 2,000 test data rows  had a toxicity score below 0.75. Nevertheless, an off-balance was noticed between the test and train set - 80.3\% entries from the test set had at least one toxic span versus a considerably higher density of 94.2\% in the train set. 

The training dataset was split into sentences while ensuring that there are no splits inside a toxic span and there are no sentences shorter than three words. Under these settings, our training dataset consists of a total of 26,589 sentences, including 10,117 records that contained toxic spans; 15\% were selected for validation. Another 2,000 entries were provided by the competition organizers for testing; the labels for this dataset were made available after the competition.

For our unsupervised training samples, we selected  20,000 random records from the Jigsaw dataset, making sure there was no overlap with the Semeval-2021 training data. Additionally, we replaced all URL-s with a special token and applied lower case on all records.

\begin{table*}[ht!]
\centering
 \begin{tabular}{p{5cm}p{1.9cm}p{2cm}} 
 \hline
  \textbf{Model} & \textbf{F1-score validation set} & \textbf{F1-score competition test set} \\
 \hline
LSTM-CRF-VAT & 75.82\% & 62.49\%  \\ 
LSTM-CRF-VAT+chars & 76.27\% & 63.65\%  \\ 
BERT-base-CRF & 79.25\% & 62.32\%  \\ 
BERT-base-CRF-VAT & 80.66\%  & 64.59\% \\ 
BERT-toxic-CRF-VAT* & \textbf{81.08}\% & 65.73\%  \\ 
BERT-news-CRF-VAT & 80.80\% & 64.57\% \\ 
BERT-news-CRF-VAT ($\gamma$=0.6) & 81.01\% & \textbf{66.13\%} \\ 
BERT-news-CRF-VAT+chars & 80.79\% & 64.57\%  \\ 
RoBERTa-large-CRF-VAT & 78.13\%  & 62.73\% \\ 
 \hline
\end{tabular}
\caption{F1-scores for predictions on the validation and test set. \\ * marks the model from the official submission.}
\label{tab:results}
\end{table*}
\subsection{Virtual Adversarial Training}

The robustness of the model in Adversarial Training is improved through examples that are close to available training data, but the model would be likely to assign a different label than the training one, thus leading to loss increase. In VAT, \citet{miyato2018virtual} adapted the adversarial training from supervised to semi-supervised settings by adding an additional loss using the Kullback–Leibler divergence between the predictions of the original data and the same data with random perturbations. Since the output distributions are compared, the information about labels is not needed for the adversarial loss: 
\begin{equation}
    L_{adv} = KL( P(\hat{y}|e,\Theta) || P(\hat{y}|e + \emph{d},\Theta))
\end{equation}
where $e$ is the embedding associated with the sample, \emph{d} the perturbation, and $\hat{y}$ is the predicted output.

True labels are required in general to compare the losses and find the worst case perturbations. However, this can be avoided  by bounding the norm of the perturbation $\delta$ to $\eta$; thus, the value of the perturbation becomes:
\begin{equation}
    \emph{d} = \argmax\limits_{\delta; ||\delta||_2<\eta}KL( P(\hat{y}|e,\Theta) || P(\hat{y}|e + \delta,\Theta))
\end{equation}

Afterwards,  we can estimate the perturbation $d$ using also the gradient $g$ and a hyperparameter $\epsilon$ for the magnitude by applying the second-order Taylor approximation and a single iteration of the power method:
\begin{equation}
    \emph{d} = \frac{g}{||g||_2}\epsilon
\end{equation}
where 
\begin{equation}
    g = \nabla_{\delta}KL( P(\hat{y}|e,\Theta) || P(\hat{y}|e + \delta,\Theta))
\end{equation}

In order to reduce the complexity and computation for the gradient, we ignore the dependency on $\Theta$. Also, the number of power iterations can be another hyperparameter for the model. The final loss function used by all models is a combination of the supervised and unsupervised adversarial loss:

\begin{equation}
    L_{total} = \gamma L_{sup} + (1-\gamma)L_{adv}
\end{equation}
where $\gamma$ is another tunable hyperparameter.

\subsection{Implementation Details}

In our experiments, pre-trained Transformer models are followed by a linear transformation of their last hidden state, and a final CRF layer. More precisely, we compare the effectiveness of several flavors of BERT models, alongside the VAT technique as follows: BERT base, a 768-dimensional model provided by Google (\textit{BERT-base-CRF-VAT}), Unitary's toxic BERT \cite{Detoxify}  (\textit{BERT-toxic-VAT}), BERT pre-trained on fake and hyperpartisan news \cite{paraschiv2020upb} (\textit{BERT-news-CRF-VAT}), and \textit{RoBERTa-large-CRF-VAT}, the equivalent of BERT-base-CRF-VAT that relies on RoBERTa instead of BERT.

In addition to these models, we experimented with enhancing the BERT-based representation with character embeddings \cite{kim2016character}. These character representations were trained on the entire Jigsaw dataset using a CNN-BiLSTM model \cite{ma2016end} with the next character prediction objective. We concatenated the obtained character-level embeddings with the aforementioned Transformer's last hidden state, and refer to this variant as \textit{BERT-news-CRF-VAT+chars}.

\begin{table*}[hbt!]
\centering
 \begin{tabular}{p{4cm}p{4cm}p{4cm}} 
 \hline
 \hline
 \Centering \textbf{Annotated sample} & \Centering \textbf{BERT-news-CRF-VAT+chars} & \Centering \textbf{BERT-toxic-CRF-VAT} \\
 \hline
 \hline
 Greedy \colorbox{lightgray}{pig} strikes again! &
 \colorbox{lightgray}{Greedy pig strikes again!} &
 \colorbox{lightgray}{Greedy pig} strikes again! \\

 \hline
What's honorable is telling \colorbox{lightgray}{idiots} that they're idiots (even if tardily) & 
 What's honorable is telling \colorbox{lightgray}{idiots} that they're \colorbox{lightgray}{idiots} (even if \colorbox{lightgray}{tardily)} &
 What's honorable is telling \colorbox{lightgray}{idiots} that they're \colorbox{lightgray}{idiots} (even if tardily)  \\
 
 \hline
 A \colorbox{lightgray}{fool} and our money are soon to be parted. & 
  \colorbox{lightgray}{A fool and our money} \colorbox{lightgray}{are soon to be parted.} &
 A \colorbox{lightgray}{fool} and our money are soon to be parted. \\
\hline
 \end{tabular}
\caption{Examples of model predictions with character embeddings (BERT-news-CRF-VAT+chars) and without (BERT-toxic-CRF-VAT), compared to the official annotations. The toxic spans are highlighted.}
\label{tab:examples2}
\end{table*}
As baseline systems, we design two methods: \textit{LSTM-CRF-VAT} with GloVe embeddings \cite{pennington2014glove} and a \textit{LSTM-CRF-VAT+chars} having character-level embeddings and VAT. In all BERT-based models, we used a maximum sequence length of 96 tokens and a sequence of 64 tokens for the LSTM baseline. Since the input words can consist of more than one token, we assign the toxicity label to a word if at least one component token is inferred as toxic.

The best hyperparameters for the BERT-base model were determined through grid search on the development set. The identified optimal values ($\epsilon=2$, $\eta=0.1$, and two power iterations) were used in all other flavors; $\gamma$ was set to 0.5 in the final loss function to balance both approaches. Furthermore, all BERT-based models were trained for one epoch in contrast with the LSTM-CRF-VAT and LSTM-CRF-VAT baselines that were trained for three epochs and four epochs, respectively.

\begin{table*}[hbt!]
\centering
 \begin{tabular}{p{7.1cm}p{7.1cm}} 
 \hline
 \hline
 \Centering \textbf{Annotated sample} & \Centering \textbf{Model predictions}\\
 \hline
 \hline
 As is Drumpf a lying gasbag. & As is Drumpf a \colorbox{lightgray}{lying gasbag}.   \\ 
 \hline
 The Liberals should listen to the economic experts and dramatically increase immigration levels (at least 500,000 per year). Stop listening to the loser racist trash. & 
 The Liberals should listen to the economic experts and dramatically increase immigration levels (at least 500,000 per year). Stop listening to the loser \colorbox{lightgray}{racist trash}.   \\
 \hline
AK Jen - Russia is no longer a communist nation. Stop spouting that - it makes you look stupid. & AK Jen - Russia is no longer a communist nation. Stop spouting that - it makes you look \colorbox{lightgray}{stupid}.   \\ 
 \hline
The "human rights" squad need a swift kick in the derriere. & The "human rights" squad need a swift \colorbox{lightgray}{kick in the derriere}.\\ 
 \hline
 Rabidly \colorbox{lightgray}{anti-Canadian troll}. & Rabidly anti-Canadian \colorbox{lightgray}{troll}.\\ 
 \hline
 Lord have Mercy on us, Trump is running \colorbox{lightgray}{amok} & Lord have Mercy on us, Trump is running amok.\\
 \hline
 \colorbox{lightgray}{You are a nut ball} & You are a \colorbox{lightgray}{nut ball} \\
 \hline
 Lives in Thailand?

So like every other racist, \colorbox{lightgray}{he’s a hypocrite} & Lives in Thailand?

So like every other racist, he’s a \colorbox{lightgray}{hypocrite} \\
 \hline
 Terry Stahlman is a worthless piece of human excrement! & Terry Stahlman is a \colorbox{lightgray}{worthless piece of human excrement!}\\

 \hline
People who are anti-immigration are weak, lying, \colorbox{lightgray}{racist whiners}.
 &  People who are anti-immigration are weak, lying, racist whiners.\\

 \hline
 Some people don’t need dope to be dopey. & Some people don’t need dope to be \colorbox{lightgray}{dopey}.\\
 \hline
 But that's what you get now for minimum wage. Increasing that to a 'livable wage' for the exact same people is stupid! & But that's what you get now for minimum wage. Increasing that to a 'livable wage' for the exact same people is \colorbox{lightgray}{stupid}! \\
 \hline
 They can't, it's jammed with an overload of their hero's \colorbox{lightgray}{excrement}.& They can't, it's jammed with an overload of their hero's excrement.\\
\hline
holy fuck you troglodytes cant even handle the simple act of meming & holy \colorbox{lightgray}{fuck you} troglodytes cant even handle the simple act of meming \\
\hline
Meanwhile Taxed is now complaining his Hyundai Santa Fe is a piece of \colorbox{lightgray}{crap} & Meanwhile Taxed is now complaining his Hyundai Santa Fe is a \colorbox{lightgray}{piece of crap} \\
\hline
Only in that \colorbox{lightgray}{sick and twisted} brain stem of yours. & Only in that sick and twisted brain stem of yours.\\
\hline
\end{tabular}
\caption{Examples from the competition test dataset of differences between the annotations and the predictions from BERT-toxic-CRF-VAT model. The toxic spans are highlighted.}
\label{tab:examples}
\end{table*}

\section{Results}
The evaluation metric for the Toxic Spans Detection task was an adapted version of the F1-score \cite{da2019fine} that takes into account the size of the overlap between prediction spans and golden labels.

Results for all developed models with the aforementioned hyperparameters (i.e., $\gamma=0.5, \epsilon=2, \eta=0.1$, and two power iterations) are presented in Table \ref{tab:results}. Since the training data had a slightly different distribution of the span density, part of our models that  performed worse on our dev set performed better on the competition test set. Adding the character embedding representation to BERT-based models did not prove to be of use in our pre-evaluation tests, but in post-evaluation, we noticed that slightly tweaking the $\gamma$ hyperparameter for the loss from 0.5 to 0.6 brought the F1-score to 66.13\%. Despite performance on the validation set was insensitive to the change in $\gamma$ between 0.5 and 0.6, the results on the test set were more than 1.5\% apart. This is mostly due to the unsupervised training that is strengthening the model's confidence on edge cases which would lower its precision. 

\begin{figure}[htp]
    \centering
    \includegraphics[width=7.8cm]{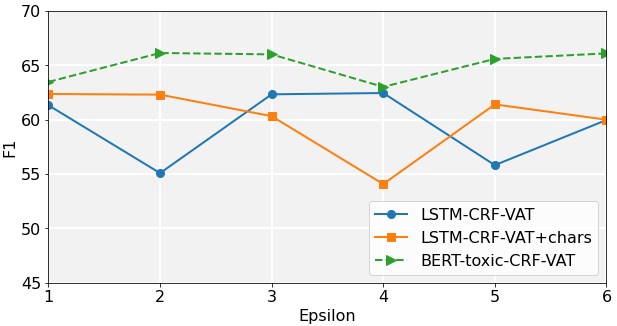}
    \caption{Impact  of the perturbation magnitude $\epsilon$ on the F1-scores for the predictions on the official test set.}
    \label{fig:epsilon}
\end{figure}

Figure \ref{fig:epsilon} introduces the influence of the perturbation magnitude $\epsilon$ on the overall performance of three models. The impact of $\epsilon$ in the adversarial training effectiveness is significant, but it is also highly dependant on the used model and can only be determined experimentally. 

Our models performed well on the detection task, learning not only common toxic expressions like "moron", "stupid", "pathetic troll", "disgusting", "hang-em high", but also obfuscated expressions like "f*cking nasty" and "b*tchy". Nonetheless, the models fail to detect more obscured words like "you don't know \textit{s***}" or "Kill this \textit{F'n W*ore} on site". All models have the tendency to over-predict toxicity by adding words to the toxic expression - for example, "What a pile of \textit{shit}" was automatically labeled as "What a \textit{pile of shit}". 

The character-level embeddings boosted the performance of the baseline LSTM-CRF-VAT model but did not improve any BERT model since it leads to detecting longer spans as toxic (see Table \ref{tab:examples2}) which in return lowers precision.

\section{Discussions and Error Analysis}

In this section, we analyze the BERT-toxic-CRF-VAT performance versus the golden label values from the competition test set. The precision and recall for our best model are 65.74\% and 85.54\%, respectively, which are indicative of a tendency to over-predict toxic spans. As we mentioned in section 3.1, even though almost all documents in the test set had a high toxicity score in the original Jigsaw dataset, many had no annotated toxic spans. Indeed, there were 295 records where our model detected a toxic span and none were labeled in the test set. Table \ref{tab:examples} includes examples of such detected spans those labeling is subjective and other detection errors. Words like "stupid", "dumb", and "crap" were assigned as toxic throughout the test data due to their high presence in the training data spans. There were also milder errors, spans that overlap with the golden labels, but the model omits part of the sequence of words. Samples like "\textit{You are a nut ball}" detected only as "You are a \textit{nut ball}" or "So like every other racist, \textit{he’s a hypocrite}" marked by the model as "So like every other racist, he’s a \textit{hypocrite}" can be perceived as likely errors even for human annotators.

\section{Conclusions and Future Work}

In this paper, several Transformer-based models (i.e., BERT and RoBERTa) were tested together with Virtual Adversarial Training to increase their robustness for identifying toxic spans from textual information. Our experiments argue that applying VAT increases performance and that domain-specific models have higher performance when compared to larger general models.

In terms of future work, we plan to experiment with self-supervised adversarial training \cite{chen2020self} to improve the robustness of our models. As we noticed in this dataset too, online users find clever ways to hide offensive and toxic expressions. Adversarial training can be effectively employed to detect these attempts and a study of its impact on offensive and hate speech classifiers is worth pursuing as follow-up leads.
\bibliographystyle{acl_natbib}
\bibliography{acl2021}


\end{document}